# Nonlinear Effects in Stiffness Modeling of Robotic Manipulators

A. Pashkevich, A. Klimchik and D. Chablat

*Abstract*— The paper focuses on the enhanced stiffness modeling of robotic manipulators by taking into account influence of the external force/torque acting upon the end point. It implements the virtual joint technique that describes the compliance of manipulator elements by a set of localized six-dimensional springs separated by rigid links and perfect joints. In contrast to the conventional formulation, which is valid for the unloaded mode and small displacements, the proposed approach implicitly assumes that the loading leads to the non-negligible changes of the manipulator posture and corresponding amendment of the Jacobian. The developed numerical technique allows computing the static equilibrium and relevant force/torque reaction of the manipulator for any given displacement of the end-effector. This enables designer detecting essentially nonlinear effects in elastic behavior of manipulator, similar to the buckling of beam elements. It is also proposed the linearization procedure that is based on the inversion of the dedicated matrix composed of the stiffness parameters of the virtual springs and the Jacobians/Hessians of the active and passive joints. The developed technique is illustrated by an application example that deals with the stiffness analysis of a parallel manipulator of the Orthoglide family.

## I. INTRODUCTION

Current trends in mechanical design of robotic manipulators are targeted at essential reduction of moving masses, in order to achieve high dynamic performances with relatively small actuators and low energy consumption. This motivates using advanced kinematical architectures (Orthoglide, Isoglide, Delta, etc.) and light-weight materials, as well as minimization of cross-sections of all critical elements. The primary constraint for such minimization is the mechanical stiffness of the manipulator, which is directly related with the robot accuracy defined by the design specifications.

In robotic literature, the manipulator stiffness is usually evaluated by a linear model, which defines the static response to the external force/torque, assuming that the compliant deflections are small and the static preloading is insignificant. However, in many practical applications (such as milling, for instance), the preloading is essential and conventional stiffness modeling techniques must be used with great caution. Moreover, for the manipulators with light-weight links, there is a potential danger of buckling phenomena that is known from general theory of elastic stability [1]. Hence, the existing stiffness modeling techniques for high-performance robotic manipulators must be revised, in order to add ability of detecting non-linear effects and avoid structural failures caused by the preloading.

The existing approaches for the manipulator stiffness modeling may be roughly divided into three main groups: the Finite Element Analysis (FEA) [2], the matrix structural analysis (SMA) [3], and the virtual joint method (VJM) that is often called the lumped modeling [4]. The most accurate of them is the Finite Element Analysis, which allows modeling links and joints with its true dimension and shape. However it is usually applied at the final design stage because of the high computational expenses required for the repeated remeshing of the complicated 3D structure over the whole workspace. The SMA also incorporates the main ideas of the FEA, but operates with rather large elements – 3D flexible beams that are presented in the manipulator structure. This leads obviously to the reduction of the computational expenses, but does not provide clear physical relations required for the parametric stiffness analysis. And finally, the VJM method is based on the expansion of the traditional rigid model by adding the virtual joints (localized springs), which describe the elastic deformations of the links, joints and actuators. The VJM technique is widely used at the pre-design stage and will be extended in this paper for the case of the preloaded manipulators.

It should be noted, that there are a number of variations and simplifications of the VJM, which differ in modeling assumptions and numerical techniques. Recent modification of this method allows to extend it to the over-constrained manipulator and to apply it at any workspace point, including the singular ones [5] [6]. Besides, to take into account real shape of the manipulator components, the stiffness parameters may be evaluated using the FEA modeling. The latter provided the FEA-accuracy throughout the whole workspace without exhaustive remeshing required for the classical FEA.

A. Pashkevich is with the Department Automatique et Productique, Ecole des Mines de Nantes, 4 rue Alfred-Kastler, Nantes 44307 France and Institut de Recherches en Communications et Cybernétiques de Nantes, 1 rue de la No, 44321 Nantes, France, (phone: 02 51 85 83 00; fax: 02 51 85 83 49; e-mail: Anatol.Pashkevich@emn.fr).

A. Klimchik is with the Department Automatique et Productique, Ecole des Mines de Nantes, 4 rue Alfred-Kastler, Nantes 44307 France and Institut de Recherches en Communications et Cybernétiques de Nantes, 1 rue de la No, 44321 Nantes, France, (e-mail: Alexandr.Klimchik@emn.fr).

D. Chablat is with Institut de Recherches en Communications et Cybernétiques de Nantes, 1 rue de la No, 44321 Nantes, France, (e-mail: Damien.Chablat@irccyn.ec-nantes.fr).

At present, there is very limited number of publication that directly addressed the problem of the stiffness modeling for preloaded manipulators. The most essential results were obtained in [7], [8] where the stiffness matrix was computed taking into account the change in the manipulator configuration due to the preloading. However, the problem of finding the corresponding loaded equilibrium was omitted, so the Jacobian and Hessian were computed in a traditional way, i.e. for the neighborhood of the unloaded equilibrium. The latter yielded essential computational simplification but also imposed essential limitations, not allowing detecting the buckling and other non-liner effects.

This paper presents a complete solution of the considered problem, taking into account influence of the external force/torque on the manipulator configuration as well as on its Jacobian and Hessian. It implements the virtual joint technique that describes the compliance of the manipulator elements by a set of localized six-dimensional springs separated by rigid links and perfect joints. The remainder of the paper is organized as follows. Section 2 defines the research problem and presents the kinetostatic model of the manipulator. In Section 3, it is proposed a numerical algorithm for computing of the loaded static equilibrium. Section 4 focuses on the stiffness matrix evaluation. And finally, Section 5 contains a numerical example that illustrates the nonlinear effects in the stiffness behavior of manipulators.

## II. PROBLEM STATEMENT

Let us consider a parallel manipulator which consists of a fixed base, several identical kinematic chains and a mobile platform. Typical examples of such kinematics (Fig. 1) are the 3-PUU translational parallel kinematic machine [9], Delta parallel robot [10], Orthoglide parallel manipulator [11] and others [12] [13].

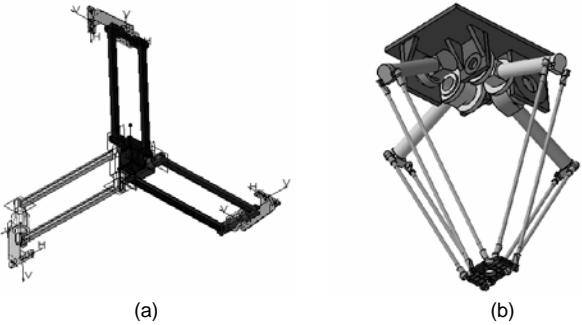

Fig. 1 Kinematics of typical parallel manipulators:
(a) Orthoglide manipulator, (b) Par 2 Fatronik manipulator

To evaluate the manipulator stiffness, let us apply the VJM method that assumes that the traditional rigid model is expended by adding virtual joints, which describe stiffness of the actuator and links. Thus, each chain of the manipulator can be described by a sequence of the following typical elements:

(a) a rigid link between the manipulator base and actuating joint described by the constant homogenous transformation matrix $\mathbf{T}_{Base}$;

(b) the 6-d.o.f. actuating joints defining three translational and three rotational actuator coordinates, which are described by the homogenous matrix function $\mathbf{T}_{Ac}(\mathbf{q}_a, \boldsymbol{\theta}_a)$ where $\mathbf{q}_a$ is the active joint coordinates and $\boldsymbol{\theta}_a$ are the virtual springs coordinate of the actuator;

(c) the 6-d.o.f. manipulator chain defining by the rigid links, virtual and passive joints coordinates, which describes by the homogenous matrix function $\mathbf{T}_{Chain}(\mathbf{q}_c, \boldsymbol{\theta}_c)$ where $\mathbf{q}_c$ and $\boldsymbol{\theta}_c$ are the vectors, which collects all passive and virtual joints coordinates of the chain respectively;

(d) a rigid link from the last link to the end-effector, described by the homogenous matrix transformation $\mathbf{T}_{Tool}$.

Hence, the end-effector position may be computed by sequential multiplication of the above homogenous matrices, so the kinematic model of a separate chain may written as

$$\mathbf{T} = \mathbf{T}_{Base} \cdot \mathbf{T}_{Ac}(\mathbf{q}_a, \boldsymbol{\theta}_a) \cdot \mathbf{T}_{Chain}(\mathbf{q}_c, \boldsymbol{\theta}_c) \cdot \mathbf{T}_{Tool} \quad (1)$$

This expression includes both traditional geometric variables (passive and active joint coordinates) and stiffness variables (virtual joint coordinates). Explicit position and orientation of the end-effector can by extracted from the matrix $\mathbf{T}$ [14], so finally the kinematic model can be rewritten as

$$t = g(\mathbf{q}, \boldsymbol{\theta}) \quad (2)$$

where $g$ is the geometry function which depends of the passive ($\mathbf{q}$) and virtual joint ($\boldsymbol{\theta}$) coordinates, the vectors $\mathbf{q} = (q_1, q_2, ..., q_n)^T$ includes all passive joint coordinates, the vector $\boldsymbol{\theta} = (\theta_1, \theta_2, ..., \theta_m)^T$ collects all virtual joint coordinates, $n$ is the number of passive joins, $m$ is the number of virtual joints.

To evaluate the manipulator ability to respond to external forces and torques, it is necessary to introduce additional equations that define the virtual joint reactions to the corresponding spring deformations. For analytical convenience, corresponding expressions may be collected in a single matrix equation

$$\boldsymbol{\tau}_{\boldsymbol{\theta}} = \mathbf{K}_{\boldsymbol{\theta}} \cdot \boldsymbol{\theta} \quad (3)$$

where $\boldsymbol{\tau}_{\boldsymbol{\theta}} = (\tau_{\theta,1}, \tau_{\theta,2}, ..., \tau_{\theta,m})^T$ is the aggregated vector of the virtual joint reactions, $\mathbf{K}_{\boldsymbol{\theta}} = diag(\mathbf{K}_{\theta,1}, \mathbf{K}_{\theta,2}, ..., \mathbf{K}_{\theta,m})$ is the aggregated spring stiffness matrix of the size m×m, and $\mathbf{K}_{\theta,i}$ is the spring stiffness matrix of the corresponding link. Similarly, one can define the aggregated vector of the passive joint reactions $\boldsymbol{\tau_q} = (\tau_{q,1}, \tau_{q,2}, ..., \tau_{q,n})^T$ but, in this case, all its components must be equal to zero $\boldsymbol{\tau_q} = \boldsymbol{0}$.

In general case, the desired stiffness model is defined by a non-liner relation

$$\mathbf{F} = f(\Delta \mathbf{t}) \quad (4)$$

that describes resistance of a mechanism to deformations $\Delta \mathbf{t}$ caused by an external force/torque $\mathbf{F}$ [1]. It should be noted that the mapping $\Delta \mathbf{t} \to \mathbf{F}$ is strictly mathematically defined and physically tractable in all cases, including under-constrained kinematics and singular configurations of the manipulator. However, the converse is not true.

In engineering practice, function $f(...)$ is usually linearized in the neighborhood of the static equilibrium $(\mathbf{q}, \boldsymbol{\theta})$ corresponding to the end-effector position $\mathbf{t}$ and external loading $\mathbf{F}$. For the unloaded mode, i.e. when $\mathbf{F} = \mathbf{0}$ and $\boldsymbol{\theta}_0 = \mathbf{0}$ the stiffness model is expressed by a simple relation

$$\mathbf{F} \approx \mathbf{K}(\mathbf{q}_0) \cdot \Delta \mathbf{t}_0 \quad (5)$$

where $\mathbf{K}$ is 6×6 "stiffness matrix" and the vector $\mathbf{q}_0 = (q_{01}, q_{02}, ..., q_{0n})^T$ defines the equilibrium configuration corresponding to the end-effector location $\mathbf{t}_0$, in accordance with the manipulator geometry.

However, for the loaded mode, stiffness model have to be defined in the neighborhood of the static equilibrium that corresponds to another manipulator configuration $(\mathbf{q}, \boldsymbol{\theta})$, which is caused by external forces $\mathbf{F}$. In this case, the stiffness model describes the relation between the increments of the force $\delta \mathbf{F}$ and the position $\delta \mathbf{t}$

$$\delta \mathbf{F} \approx \mathbf{K}(\mathbf{q}, \mathbf{t}) \cdot \delta \mathbf{t} \quad (6)$$

where $\mathbf{q} = \mathbf{q}_0 + \Delta \mathbf{q}$ and $\boldsymbol{\theta} = \boldsymbol{\theta}_0 + \Delta \boldsymbol{\theta}$ denote the new position of the manipulator, $\Delta \mathbf{q}$ and $\Delta \boldsymbol{\theta}$ are the deviations of the passive joint and virtual spring coordinates.

Hence, the problem of the stiffness modeling in the loaded mode may be divided into two sequential subtasks: (i) finding the static equilibrium for the loaded configuration; and (ii) linearization of relevant force/position relations in the neighborhood of this equilibrium. Let us consider these two sub-problems consequently.

### III. STATIC EQUILIBRIUM FOR THE LOADED MODE

Let us assume that, due to the external force F, the end-effector of the manipulator is relocated from the initial (unloaded) position $\mathbf{t}_0 = P(\mathbf{q}_0, \boldsymbol{\theta}_0)$ to a new position $\mathbf{t} = P(\mathbf{q}, \boldsymbol{\theta})$, which satisfies the condition of the mechanical equilibrium. Here $\mathbf{q}_0$ is computed via the inverse kinematic and $\boldsymbol{\theta}_0$ is equal to zero (since there are no external loading in the springs), $\mathbf{q}, \boldsymbol{\theta}$ are passive and virtual joint coordinate in the loaded mode respectively. For rather small displacement $\Delta \mathbf{t} = \mathbf{t} - \mathbf{t}_0$, a new position of the end-effector $\mathbf{t} = P(\mathbf{q}_0 + \Delta \mathbf{q}, \boldsymbol{\theta}_0 + \Delta \boldsymbol{\theta})$ may be expressed as

$$\mathbf{t} = \mathbf{t}_0 + \mathbf{J}_\theta \cdot \Delta \boldsymbol{\theta} + \mathbf{J}_q \cdot \Delta \mathbf{q} \quad (7)$$

where $\mathbf{J}_\theta$ and $\mathbf{J}_q$ are the kinematic Jacobians with respect to the coordinates $\boldsymbol{\theta}, \mathbf{q}$, which may be computed from (1) analytically or semi-analytically, using the factorization technique [6]. However, in general case, the stiffness model is highly non-linear and computing $(\mathbf{q}, \boldsymbol{\theta})$ requires some additional efforts.

For computational reasons, let us consider the dual problem that deals with determining the external force F and the manipulator configuration $(\mathbf{q}, \boldsymbol{\theta})$ that correspond to the output position $\mathbf{t}$.

Let us assume that the joints give small, arbitrary virtual displacements $\Delta \boldsymbol{\theta}$ in the equilibrium neighborhood. According to the principle of virtual displacements, the virtual work of the external force $\mathbf{F}$ applied to the end-effector along the corresponding displacement $\Delta \mathbf{t} = \mathbf{J}_\theta \cdot \Delta \boldsymbol{\theta} + \mathbf{J}_q \cdot \Delta \mathbf{q}$ is equal to the sum $(\mathbf{F}^T \mathbf{J}_\theta) \cdot \Delta \boldsymbol{\theta} + (\mathbf{F}^T \mathbf{J}_q) \cdot \Delta \mathbf{q}$. Since the passive joints do not produce the force/torque reactions, the virtual work includes only one component $-\boldsymbol{\tau}_\theta^T \cdot \Delta \boldsymbol{\theta}$ (the minus sign takes into account the force-displacement directions for the virtual spring). In the static equilibrium, the total virtual work of all forces is equal to zero for any virtual displacement, therefore the equilibrium conditions may be written as

$$\begin{aligned} \mathbf{J}_\theta^T \cdot \mathbf{F} &= \boldsymbol{\tau}_\theta; \\ \mathbf{J}_q^T \cdot \mathbf{F} &= \mathbf{0} \end{aligned} \quad (8)$$

Taking into account (3), the latter can be rewritten as

$$\begin{aligned} \mathbf{F} \cdot \mathbf{J}_\theta^T &= \mathbf{K}_\theta \cdot \boldsymbol{\theta}; \\ \mathbf{F} \cdot \mathbf{J}_q^T &= \mathbf{0} \end{aligned} \quad (9)$$

It is evident that there is no general method for analytical solution of this system and it is required to apply numerical techniques. To derive the numerical algorithm, let us linearize the kinematic equation in the neighborhood of the current position $(\mathbf{q}_i, \boldsymbol{\theta}_i)$

$$\mathbf{t} = P(\mathbf{q}_i, \boldsymbol{\theta}_i) + \mathbf{J}_q(\mathbf{q}_i, \boldsymbol{\theta}_i) \cdot (\mathbf{q}_{i+1} - \mathbf{q}_i) + \mathbf{J}_\theta(\mathbf{q}_i, \boldsymbol{\theta}_i) \cdot (\boldsymbol{\theta}_{i+1} - \boldsymbol{\theta}_i) \quad (10)$$

and rewrite the static equilibrium equations as

$$\mathbf{J}_\theta^T(\mathbf{q}_i, \mathbf{F}_i) \cdot \mathbf{K}_\theta = \theta_{i+1};$$
$$\mathbf{J}_q^T(\mathbf{q}_i, \mathbf{F}_i) \cdot \theta_{i+1} = \quad (11)$$

This leads to a linear algebraic system of equations with respect to $(\mathbf{q}_{i+1}, \mathbf{F}_{i+1}, \theta_{i+1})$

$$\mathbf{J}_q(\theta_i, \mathbf{q}_i) \cdot \mathbf{q}_{i+1} + \mathbf{J}_q(\mathbf{q}_i, \theta_i) \cdot \theta_{i+1} = $$
$$= \mathbf{t} - \mathbf{g}(\mathbf{q}_i, \mathbf{J}_i) \cdot \mathbf{q}_{i} - \mathbf{g}(\mathbf{q}_i, \mathbf{q}_i) \cdot \mathbf{J}_i \cdot \mathbf{q}_i \cdot \theta(\mathbf{q}_i, \theta_i) \cdot \theta_i \quad (12)$$
$$-\mathbf{K}_\theta \cdot \theta_{i+1} + \mathbf{J}_\theta^T(\theta_i, \mathbf{F}_i) \cdot \mathbf{F}_{i+1} = 0$$
$$\mathbf{J}_q^T(\mathbf{q}_i, \mathbf{F}_i) \cdot \theta_{i+1} = $$

which gives the following iterative scheme

$$\begin{bmatrix} \mathbf{F}_{i+1} \\ \mathbf{q}_{i+1} \end{bmatrix} = \begin{bmatrix} \mathbf{J}_\theta(\mathbf{q}_i, \mathbf{K}_i) \mathbf{J}_\theta^{-1} \mathbf{q}_\theta^T \theta_i, & \mathbf{J}_q \mathbf{q}_q(\theta_i, _i) \\ \mathbf{J}_q^T(\mathbf{q}_i, _i) & 0 \end{bmatrix}^{-1} \times$$
$$\times \begin{bmatrix} \mathbf{t} - \mathbf{g}(\mathbf{q}_i, \mathbf{J}_i) \cdot \mathbf{q}_i \cdot \mathbf{g}(\mathbf{q}, _i) \mathbf{J}_i \cdot \mathbf{q}_i \cdot \theta(\theta, _i)_i \\ 0 \end{bmatrix} \quad (13)$$
$$\theta_{i+1} = \mathbf{K}_\theta^{-1} \cdot \mathbf{J}_\theta^T(\mathbf{q}_i, \theta_i) \cdot \mathbf{F}_{i+1}$$

where the starting point $(\mathbf{q}, \theta, _0)$ can be chosen using the non-loaded configuration, and computed via the inverse kinematics.

As follows from computational experiments, for typical values of deformations the proposed iterative algorithm possesses rather good convergence (3-5 iterations are usually enough). However, in the case of buckling or in the area of multiple equilibriums, the problem of convergence becomes rather critical and highly depends on the initial guess. To overcome this problem, the value of the joint variables $(\theta_i, \mathbf{q}_i)$ computed at each iterations were disturbed by adding small random noise. Further enhancement of this algorithm may be based on the full-scale Newton-Raphson technique (i.e. linearization of the static equilibrium equations in addition to the kinematic one), this obviously increases computational expenses but potentially improves convergence.

## IV. Stiffness Model For The Loaded Mode

After the static equilibrium corresponding to the external loading is found, the force-displacement relations may be linearized. To compute the desired stiffness matrix, let us assume that the manipulator was moved from the configuration $(\mathbf{F}, \theta, \mathbf{t}, )$ to the configuration $(\mathbf{F} + \delta \mathbf{F}, \theta, \mathbf{q} + \delta \mathbf{q}, \mathbf{t} + \delta , +\delta )$ and both of the satisfy the equilibriums equations, i.e.

$$\mathbf{F} \cdot \mathbf{J}_\theta^T = \mathbf{K}_\theta \cdot \quad ;$$
$$\mathbf{F} \cdot \mathbf{J}_q^T = 0 \quad (14)$$

and

$$(\mathbf{F} + \delta \mathbf{F})(\mathbf{J}_\theta + \delta \mathbf{J}_\theta)^T = \mathbf{K}_\theta \cdot ( + \delta );$$
$$(\mathbf{F} + \delta \mathbf{F}) \cdot (\mathbf{J}_q + \delta \mathbf{J}_q)^T = 0 \quad (15)$$

where $\delta \mathbf{J}_q(\mathbf{q}, )$ and $\delta \mathbf{J}_\theta(\mathbf{q}, )$ are the differentials of the Jacobians due to changes in $(\mathbf{q}, \theta)$.

Let us also linearize the geometric model (5) in the neighborhood of $(\mathbf{q}, \theta)$

$$\delta \mathbf{t} = \mathbf{J}_\theta(\mathbf{q}, \theta) \cdot \delta \theta + \mathbf{J}_q(\mathbf{q}, \theta) \cdot \delta \mathbf{q}, \quad (16)$$

After relevant transformation and neglecting high-order small terms, equations (14), (15) may be rewritten as

$$\mathbf{J}_\theta^T(\mathbf{q}, \mathbf{F}) \delta \mathbf{F} + \mathbf{H}_{\theta q}^F \delta \mathbf{q} + \mathbf{H}_{\theta \theta}^F \delta \theta - \mathbf{K}_\theta \delta \theta = $$
$$\mathbf{J}_q^T(\mathbf{q}, \mathbf{F}) \delta \mathbf{F} + \mathbf{H}_{qq}^F \delta \mathbf{q} + \mathbf{H}_{q\theta}^F \delta \theta = 0 \quad (17)$$

where $\mathbf{H}_{qq}^F, \mathbf{H}_{\theta q}^F, \mathbf{H}_{q\theta}^F, \mathbf{H}_{\theta\theta}^F$ are the Hessian matrices of the scalar function $\mathbf{F}^T g(\mathbf{q}, \theta)$.

This allows to apply substitution for $\delta \theta$ and to obtain system of two matrix equations with unknowns $\delta \mathbf{F}$ and $\delta \mathbf{q}$

$$\begin{bmatrix} \mathbf{J}_\theta \cdot \mathbf{k}_\theta^F \cdot \mathbf{J}_\theta^T & \mathbf{J}_q + \mathbf{J}_\theta \cdot \mathbf{k}_\theta^F \cdot \mathbf{H}_{\theta q}^F \\ \mathbf{J}_q^T + \mathbf{H}_{\theta q}^F \cdot \mathbf{k}^F \cdot \mathbf{J}_{qq}^T & \mathbf{H}_{q\theta}^F + \mathbf{H}_{\theta q}^F \cdot \mathbf{k}_\theta^F \cdot \mathbf{H}^F \end{bmatrix} \cdot \begin{bmatrix} \delta \mathbf{F} \\ \delta \mathbf{q} \end{bmatrix} = \begin{bmatrix} \delta \mathbf{t} \\ 0 \end{bmatrix}, \quad (18)$$

which determine stiffness model for the case of the loaded equilibrium. Here $\mathbf{k}_\theta^F = (\mathbf{K}_\theta - \mathbf{H}_{\theta\theta}^F)^{-1}$.

Therefore, for a separate kinematic chain, the desired stiffness matrix $\mathbf{K}^F$ defining the displacement-to-force mapping (4) in the neighborhood of the loaded configuration can be computed by direct inversion of the matrix in the left-hand side of (18) and extracting from it the left-upper 6×6 sub-matrix.

Let us note that the matrix (18) can be computed and inverted for any configuration (including singular ones). Besides, the proposed technique takes into account both elastic deformations in the virtual springs and unrestricted kinematics motions due to the passive joints. In the case of multi-chain manipulator, the desired matrix can be computed by simple summation $\mathbf{K}^F = \sum_{i=1}^n \mathbf{K}_i^F$, where $\mathbf{K}_i^F$ corresponds to the i-th chain. This follows from the superposition principle, since the total external force corresponding to the end-effector displacement $\delta \mathbf{t}$ (the same for each kinematic chains) can be expressed as the sum of the partial forces.

## V. ILLUSTRATIVE EXAMPLE

### A. Kinetostatic model

Let us illustrate the proposed technique by stiffness analysis of a translational manipulator of the Orthoglide family presented in Fig.1. For this manipulator, each kinematic chain consists of a foot, a kinematic parallelogram with two axes and two bars, and an end-effector. (Fig. 2).

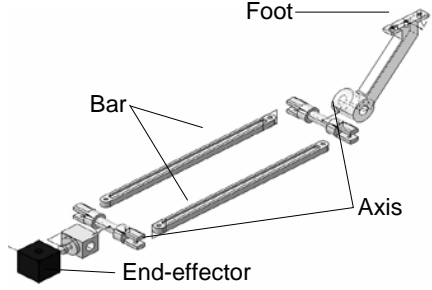

Fig 2 Chain of the Orthoglide manipulator

As follows from a separate study, the rigidity of the parallelogram axes is high compared to the bar and the foot. For the remaining elements, the compliance matrices (i.e. the inverses of the corresponding stiffness matrices) were identified using the FEA-based methodology and special accuracy improvement tools proposed by the authors [15]. Numerical values for these matrices are

$$k_{Foot} = \begin{bmatrix} 28 \cdot 10^{-5} & -33 \cdot 10^{-5} & 0 & 0 & 0 & -40 \cdot 10^{-7} \\ -33 \cdot 10^{-5} & 41 \cdot 10^{-5} & 0 & 0 & 0 & 54 \cdot 10^{-7} \\ 0 & 0 & 19 \cdot 10^{-4} & 11 \cdot 10^{-6} & -15 \cdot 10^{-6} & 0 \\ 0 & 0 & 11 \cdot 10^{-6} & 23 \cdot 10^{-8} & 0 & 0 \\ 0 & 0 & -15 \cdot 10^{-6} & 0 & 23 \cdot 10^{-8} & 0 \\ -40 \cdot 10^{-7} & 54 \cdot 10^{-7} & 0 & 0 & 0 & 84 \cdot 10^{-9} \end{bmatrix} \quad (19)$$

$$k_{Bar} = \begin{bmatrix} 46 \cdot 10^{-6} & 0 & 0 & 0 & 0 & 0 \\ 0 & 23 \cdot 10^{-2} & 0 & 0 & 0 & 11 \cdot 10^{-5} \\ 0 & 0 & 51 \cdot 10^{-3} & 0 & -24 \cdot 10^{-5} & 0 \\ 0 & 0 & 0 & 29 \cdot 10^{-6} & 0 & 0 \\ 0 & 0 & -24 \cdot 10^{-5} & 0 & 15 \cdot 10^{-7} & 0 \\ 0 & 11 \cdot 10^{-4} & 0 & 0 & 0 & 72 \cdot 10^{-7} \end{bmatrix} \quad (20)$$

For this analysis, the kinematic parallelogram was replaced by a bar element with double stiffness and there were considered several typical postures presented in Fig 3.

For such approximation, the kinematic model of the manipulator chain is expressed as

$$\mathbf{T} = \mathbf{T}_{base} \cdot \mathbf{T}_x(\theta_a) \cdot \mathbf{V}(\boldsymbol{\theta}_{Foot}) \cdot \mathbf{R}_y(q_1) \cdot \mathbf{R}_z(q_2) \times \\ \times \mathbf{T}_x(\theta_a) \cdot \mathbf{V}(\boldsymbol{\theta}_{Link}) \cdot \mathbf{R}_z(q_3) \cdot \mathbf{R}_y(q_4) \cdot \mathbf{T}_{Tool} \quad (21)$$

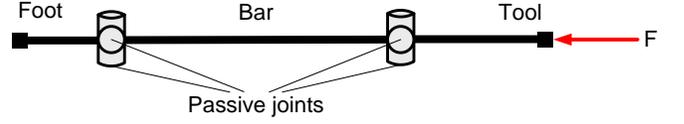

(a) Posture A

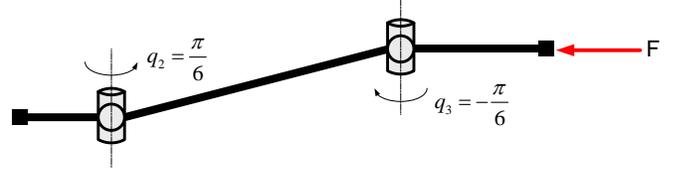

(b) Posture B

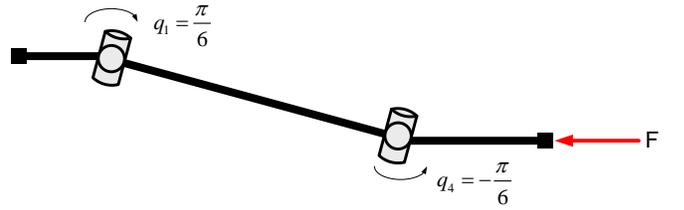

(c) Posture C

Fig 3 Typical postures of the Orthoglide kinematic chain

where $\mathbf{T}_{base}, \mathbf{T}_{Tool}$ are constant transformations matrices (the matrix $\mathbf{T}_{base}$ includes also the "foot" transformation); $\theta_a, \boldsymbol{\theta}_{Foot}, \boldsymbol{\theta}_{Link}$ are the virtual joint coordinates of the actuator, the "foot" and the "link" respectively;. $\mathbf{V}(...)$ is the homogeneous matrix-function of the virtual springs, which depends on six variables and can be described via the multiplication of elementary transformations as

$$\mathbf{V}(\theta_x, \theta_y, \theta_z, \theta_{\varphi x}, \theta_{\varphi y}, \theta_{\varphi z}) = \\ = \mathbf{T}_x(\theta_x) \cdot \mathbf{T}_y(\theta_y) \cdot \mathbf{T}_z(\theta_z) \cdot \mathbf{R}_x(\theta_{\varphi x}) \cdot \mathbf{R}_y(\theta_{\varphi y}) \cdot \mathbf{R}_z(\theta_{\varphi z}) \quad (22)$$

where $\mathbf{T}_x, \mathbf{T}_y, \mathbf{T}_z, \mathbf{R}_x, \mathbf{R}_y, \mathbf{R}_z$ are elementary homogeneous transformation matrices. For this case study, the Jacobian and Hessian matrices were computed semi-analytically, via differentiation of the model (21).

### B. Stiffness modeling

The stiffness modeling experiments were carried out for four typical postures presented in Fig 3 and TABLE I. During the stiffness modeling, the end-effector of the kinematic chain was displaced in the range between 0 and 4 mm with the step 0.001 mm, starting from the unloaded configuration. The static equilibrium and corresponding force were determined for each displacement using the iterative algorithm presented in Section 3. Besides, the stiffness matrix was computed for each case. The stiffness model was computed sequentially from small to high displacement, using previous state as a starting one for the next step. The force-displacement relationship for each posture present on the Fig 4

TABLE I
CONFIGURATIONS OF THE MANIPULATOR CHAIN

| Configuration | $q_1$ | $q_2$ | $q_3$ | $q_4$ |
|---|---|---|---|---|
| Pos. A (Fig 3, a) | 0 | 0 | 0 | 0 |
| Pos. B (Fig 3, b) | 0 | $\pi/6$ | $-\pi/6$ | 0 |
| Pos. C (Fig 3, c) | $\pi/6$ | 0 | 0 | $-\pi/6$ |
| Pos. D | $\pi/6$ | $\pi/6$ | $-\pi/6$ | $-\pi/6$ |

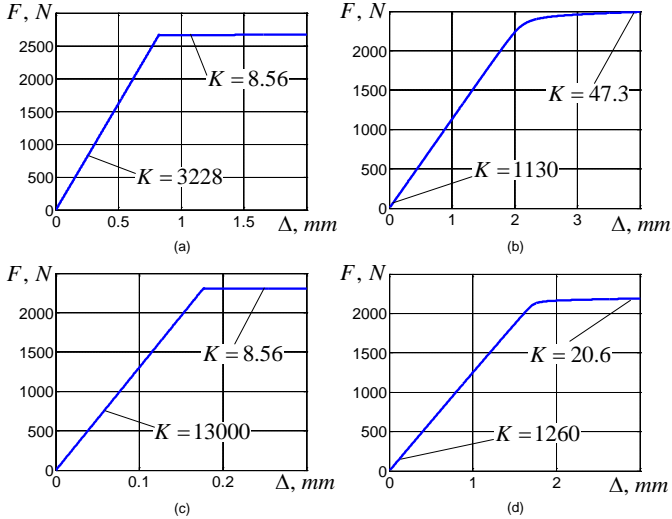

Fig 4 Force-displacement relationship of the Orthoglide chain:
a) Pos. A; b) Pos. B; c) Pos. C; d) Pos. D;

The obtained results show that the stiffness model of the considered kinematic chain is significantly nonlinear and essentially depends on both the posture of the manipulator chain and the external force. In particular, applying the force about 2 kN leads to the buckling effect and considerable reduction of the stiffness (by 50…100 times, see TABLE I). After the buckling, even insignificant increase of the external force causes very essential deflections of the kinematic chain and corresponding positing errors.

Hence, the developed technique allowed detecting some uncommon behavior of the robotic manipulators which was previously not reported in robotic literature. These phenomena are directly related to the robot accuracy must be obviously taken into account during design and analysis.

ACKNOWLEDGMENT

The work presented in this paper was partially funded by the Region "Pays de la Loire", France and by the EU commission (project NEXT).

TABLE II
INFLUENCE OF THE EXTERNAL LOADING ON THE STIFFNESS OF THE MANIPULATOR CHAIN

| Posture | $K_0$, N/m | Buckling | | | | Large deformations | | |
|---|---|---|---|---|---|---|---|---|
| | | $K_1$, N/m | $F_{cr}$, N | $\Delta_{cr}$, mm | $K_2$, N/m | $F_1$, N | $\Delta_1$, mm | $K_3$, N/m |
| Posture A | 3228 | 3228 | 2661 | 0.82 | 8.56 | 2672 | 2.00 | 8.56 |
| Posture B | 1130 | 1105 | 1920 | 1.7 | 159 | 2494 | 4.00 | 27.7 |
| Posture C | 13056 | 13090 | 2300 | 0.18 | 8.56 | 2305 | 0.30 | 8.56 |
| Posture D | 1260 | 1170 | 2000 | 1.61 | 52.73 | 2191 | 3.00 | 20.6 |

$K_0$ is the stiffness for the unloaded mode, $K_1$ is the stiffness before the buckling, $K_2$ is the stiffness after the buckling, $K_3$ is the stiffness for the "large" deformations, $F_{cr}$ is the critical force for the buckling, $F_1$ is the force for the "large" deformations, $\Delta_{cr}$ is the deformation in the buckling mode, $\Delta_1$ is the "large" deformation value.